\documentclass[sigconf]{acmart}

\AtBeginDocument{%
  }

\setcopyright{acmlicensed}
\copyrightyear{2018}
\acmYear{2018}
\acmDOI{XXXXXXX.XXXXXXX}

\acmConference[Conference acronym 'XX]{Make sure to enter the correct
  conference title from your rights confirmation emai}{June 03--05,
  2018}{Woodstock, NY}
\acmISBN{978-1-4503-XXXX-X/18/06}

\usepackage{bm}
\usepackage{multirow}
\usepackage{color, colortbl}
\definecolor{Gray}{gray}{0.9}

\begin{document}
%\title{Using Large Language Models for Assistance in Dispute Resolution}
%\title{DRAssist: Using Large Language Models for Assistance in Dispute Resolution}
\title{DRAssist: Dispute Resolution Assistance\\ using Large Language Models}
\author{Sachin Pawar, Manoj Apte, Girish K. Palshikar, Basit Ali, Nitin Ramrakhiyani}
%\email{author1@corporation.com}
%\author{Author2}
%\email{author2@corporation.com}
%\author{Author3}
%\author{Author4}
%\email{{author1,author2,author3,author4}@corporation.com}
\email{{sachin7.p,manoj.apte,gk.palshikar,ali.basit,nitin.ramrakhiyani}@tcs.com}
\affiliation{
\institution{TCS Research, Tata Consultancy Services Limited}
\city{Pune}
%\state{XYZ}
\country{India}
}
\renewcommand{\shortauthors}{Pawar et al.}

\begin{abstract}
  %\textcolor{red}{TODO: to be completed.}
  Disputes between two parties occur in almost all domains such as taxation, insurance, banking, healthcare, etc. The disputes are generally resolved in a specific forum (e.g., consumer court) where facts are presented, points of disagreement are discussed, arguments as well as specific demands of the parties are heard, and finally a human judge resolves the dispute by often favouring one of the two parties. In this paper, we explore the use of large language models (LLMs) as assistants for the human judge to resolve such disputes, as part of our {\bf DRAssist} system. We focus on disputes from two specific domains -- automobile insurance and domain name disputes. DRAssist identifies certain key structural elements (e.g., facts, aspects or disagreement, arguments) of the disputes and summarizes the unstructured dispute descriptions to produce a structured summary for each dispute. We then explore multiple prompting strategies with multiple LLMs for their ability to assist in resolving the disputes in these domains. 
  In DRAssist, these LLMs are prompted to produce the resolution output at three different levels -- (i) identifying an overall stronger party in a dispute, (ii) decide whether each specific demand of each contesting party can be accepted or not, (iii) evaluate whether each argument by each contesting party is strong or weak. We evaluate the performance of LLMs on all these tasks by comparing them with relevant baselines using suitable evaluation metrics.
\end{abstract}




\maketitle

\section{Introduction}

A {\em dispute} is rooted in some kind of disagreement, conflict, dissatisfaction, violations, grievance or difference of opinion between two or more {\em parties}. Disputes occur in almost all government and business domains, such as taxation, insurance, banking, stock markets, manufacturing, healthcare etc. A dispute is often {\em pre-legal} i.e., it is raised in a non-judicial forum, such as an organization's internal unit or sometimes, with an authorized external body such as a tribunal or commission; these may or may not involve legal professionals such as lawyers. %Most disputes, if not resolved in the pre-legal stage, end up as lawsuits or litigations in legal institutions such as courts.
Unresolved disputes in the pre-legal stage, end up as litigations in legal institutions such as courts.

A common form of pre-legal disputes is where one party is an {\em entity} (which is typically an organization such as a manufacturing company, a service provider company, or a government agency e.g., for taxation) and the other party is its {\em customer(s)}. Here, the entity is usually one of the parties in the disputes. The other form of disputes does not necessarily involve such a common entity. Disputes of the first form are widespread in modern businesses and government organizations. Examples: tax disputes (entity is the government taxation authority), mobile phone quality disputes (entity is the mobile phone manufacturer), service disputes (entity is a bank), insurance claim disputes (entity is the insurance company) etc. Many issues, such as violent crimes, are obviously {\em not} included under disputes. %In this paper, we use automobile insurance disputes (first form) and domain name disputes (second form). 
In this paper, we consider both forms of disputes -- automobile insurance disputes (first form) and domain name disputes (second form). 

{\em Resolution} of a dispute involves a {\em decision}, and/or suggested {\em actions / consequences}, which are binding on both parties. In case the decision is not acceptable to one or both parties, then the dispute is escalated to a higher authority, such as a court of law or an appellate. In the pre-legal stage, dispute resolution is often done through an {\em adjudicative process} by a judge or an arbitrator, or sometimes in a {\em consensual process}, such as mediation, conciliation, or negotiation, in which the two parties try to reach a mutually agreeable resolution. Dispute resolution is effort-intensive, time-consuming, tedious and demands knowledge of the relevant laws, domain-specific documents (e.g., insurance policies) or past resolutions. Unresolved disputes, or disputes that escalate into litigation damage the reputation and goodwill of the entity, lead to customer dissatisfaction and may cause financial or material losses. 

Disputes often involve {\em demands} by both contesting parties; e.g., a vehicle owner may {\it demand a larger monetary compensation for total loss of the vehicle}, whereas the insurance company may {\it deny that the vehicle is a total loss} and may be willing to pay only for the repairs. Resolution of a dispute can be viewed in different ways. In {\em hard resolution}, the winning party is identified, punishments/penalties are assigned to the non-winning party, other consequences (e.g., costs, damages) are declared, and all (or most) demands of the winning party - and none of the other party - are accepted, perhaps with some modifications. In {\em mixed or partial resolution}, there is no clearly identified winning party, and the demands of both parties may be partially or fully accepted. In this paper, our focus is not on designing an automated dispute resolution system, but rather designing a system that provides {\em assistance} to the human expert in preparing an informed resolution of the given dispute.  

%Figure~\ref{fig_ex} shows an example dispute\footnote{https://taxguru.in/goods-and-service-tax/maruti-case-no-reduction-gst-rate-section-171-attracted.html} related to the Goods and Services Tax (GST) in India~\cite{RM16}, which is similar to the VAT in many countries\footnote{ https://www.ey.com/gl/en/services/tax/vat--gst-and-other-sales-taxes/ey-ch2-1-avoid-vat-gst-errors-and-disputes}; the relevant law is The Central Goods and Services Tax Act (CGST), 2017. 

\begin{table}[]\footnotesize
\caption{Structured summary of an auto insurance dispute.}\label{tabStructuredSummary}
\begin{tabular}{p{0.97\columnwidth}}
\hline
\textbf{Facts Agreed by Both Parties}: The vehicle was purchased on 29.11.2002. The vehicle was insured on 19.12.2002. The vehicle was reported stolen on 23.12.2002, with the complainant claiming it was stolen from Bodhgaya. The insurance company contends that the vehicle was stolen from Kaohhar Kainur on 18.12.2002.\\
\hline
\textbf{Aspects on which the parties disagree}:\newline
Date of Theft: The insured party claims the vehicle was stolen on 23.12.2002. The insurance company asserts that the vehicle was stolen on 18.12.2002.\newline
Place of Theft: The insured party contends that the theft occurred in Bodhgaya. The insurance company claims the vehicle was stolen from Kaohhar Kainur.\newline
Timing of Insurance Coverage: The insurance company argues that the theft occurred before the insurance was taken, while the insured party maintains that the theft occurred after the insurance coverage was in effect.\\
\hline
\textbf{Demands of the insurance company}: Denial of the insurance claim. Assertion that the policy was fraudulently obtained. Repudiation of the claim based on the fraudulent nature of the policy acquisition.\\
\hline
\textbf{Demands of the insured party}: Payment of the insured amount for the stolen vehicle. Interest on the insured amount at the rate of 10\% per annum.\\
\hline
\textbf{Arguments of the insurance company}: The insurance company asserts that the vehicle in question was stolen prior to the issuance of the insurance policy, specifically on 18.12.2002. The insured party allegedly misrepresented the vehicle during the inspection process by presenting a different vehicle, leading to the fraudulent acquisition of the insurance policy on 19.12.2002. The insurance company has acknowledged negligence on the part of its functionary, who failed to conduct a proper physical inspection of the vehicle, and has taken departmental action against this individual for their oversight. Based on these grounds, the insurance company contests the validity of the insurance policy, claiming it was obtained under fraudulent circumstances.\\
\hline
\textbf{Arguments of the insured party}: The insured party purchased the vehicle on 29.11.2002, with the assurance from the dealer that the insurance policy would be obtained on his behalf. Due to delays from the dealer in securing the insurance policy, the insured party took the initiative to obtain the policy himself on 19.12.2002. The insured party asserts that the vehicle was shown to the functionary of the insurance company at the time the policy was obtained. The vehicle was stolen from Bodhgaya on 23.12.2002, after the insurance policy was in effect. The insured party claims that the police investigation into the theft confirmed the incident, and the police report has been accepted by the competent court, supporting the validity of the insured party's claims regarding the theft.\\
\hline
\textbf{Prior disputes referred with short summary}: No Specific Prior Cases Mentioned\\
\hline
\textbf{Statutes or policy terms and conditions referred with short summary}: No Specific Statutes Referenced. There are no specific policy terms or conditions mentioned by either party.\\
\hline
\textbf{Decision by District Commission}: The Commission directed the complainant to seek remedy in the competent civil court.\\
\hline
\textbf{Decision by State Commission}: The Commission directed the insurance company to pay the insured amount for the stolen vehicle. The payment is to include interest at the rate of 10\% per annum.\\
\hline
\textbf{Final decision by the National Commission for each demand of both parties}: The insurance company's revision petition has been dismissed. The National Commission upholds the order of the State Commission requiring the insurance company to pay the insured amount for the stolen vehicle. The insurance company is directed to pay the insured amount along with interest at the rate of 10\% per annum. The insurance company is not required to pay any additional amounts beyond the insured amount and interest.\\
\hline
\textbf{Justification / rationale for the final decision}: The police investigation did not find any discrepancies regarding the place and date of the incident as claimed by the complainant. The police report, which confirmed the occurrence of the theft, was duly accepted by the competent judicial court. The National Commission emphasized that there were no grounds to doubt the validity of the complainant's claims based on the accepted police report. Consequently, the National Commission found no reason to overturn the State Commission's decision, which relied on the police report and its acceptance by the court.\\
\hline
\textbf{Winning party}: the insured party.\\*
\hline
\end{tabular}
\end{table}
\begin{figure}[htbp]
\centering
\includegraphics[width=\columnwidth,height=0.3\columnwidth]{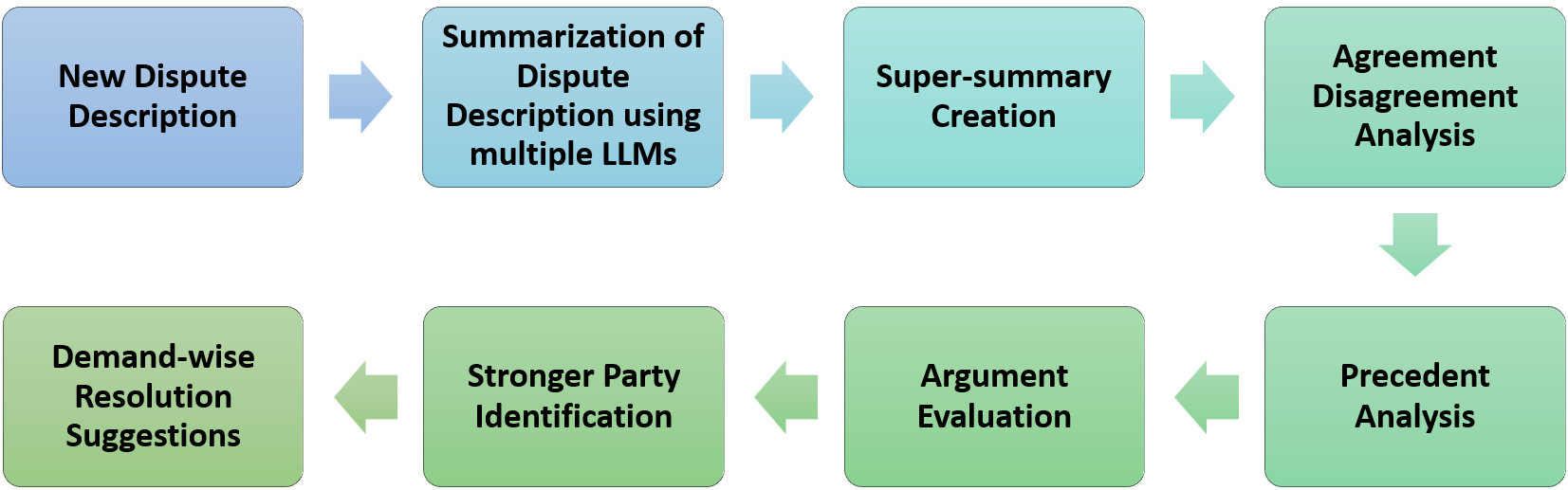}
\caption{Workflow in dispute resolution assistance system.}
\label{fig1}
\end{figure}

As an example, consider a dispute between a truck owner and an insurance company\footnote{https://e-jagriti.gov.in/case-history-case-status go to tab "CASE HISTORY" and search with Commission Name = NCDRC and Case Number = RP/2142/2016}. Table~\ref{tabStructuredSummary} shows a structured summary of this dispute generated using an LLM (discussed later in Section~\ref{secStructuredSummary}).   

In this paper, we present our {\bf DRAssist} system where we demonstrate the use of {\em Large Language Models} (LLMs) as {\em assistants} to human experts (arbitrators, adjudicators, lawyers etc.) for {\em dispute resolution} (DR). Figure~\ref{fig1} shows the pipeline of tasks in our DR workflow. DRAssist's primary goal is {\em not} to provide an automated one-shot DR as such. Instead, in the spirit of a consensual process of DR, we view DR as a multi-step process, and provide AI-based assistance in each step. Major steps in this consensual process are: 
% \begin{enumerate}
%     %\item Comparison of Facts related to the dispute; identification of facts over which the two parties disagree.
%     \item Comparison of legal and domain-specific arguments presented by each side
%     \item Comparison of precedents and prior disputes cited by both sides as part of their arguments
%     \item Identification of the {\em stronger party} by taking into account the above 3 steps, along with a human-understandable {\em justification} for this identification. Note that the identified stronger party is only as an assistance to the human decision-maker and is {\em not} to be thought of as winning party.
%     \item Resolution of demands made by both parties. As a simple model of this {\em demand resolution}, each demand is predicted as {\small {\sf ACCEPTED}} or {\small {\sf REJECTED}}, along with a justification for each predicted label. Again, this is not to be viewed as a final decision, but only as an assistance to the human decision-maker. A consistency requirement on demand resolution needs to be enforced, viz., that at most one of a pair of conflicting demands (one from each contesting parties) can be of {\small {\sf ACCEPTED}}. This soft approach keeps open the possibility that some demands of one party and some demands of the other party may be {\small {\sf ACCEPTED}}, as discussed above. 
% \end{enumerate}

\textit{(1)} Analysis of facts related to the dispute -- the aspects on which both the parties agree and the aspects where there is disagreement.

\textit{(2)} Analysis of precedents and prior disputes cited by both sides as part of their arguments
    
\textit{(3)} Comparison of legal and domain-specific arguments presented by each side to evaluate each argument as {\it strong} or {\it weak}.

\textit{(4)} Identification of the {\em stronger party} by taking into account the above 3 steps, along with a human-understandable {\em justification} for this identification. Note that the identified stronger party is shown to the human decision-maker only as an assistance and is {\em not} to be thought of as the winning party.
    
\textit{(5)} Resolution of demands made by both parties. As a simple model of this {\em demand resolution}, each demand is predicted as {\small {\sf ACCEPTED}} or {\small {\sf REJECTED}}, along with a justification for each predicted label. Again, this is not to be viewed as a final decision, but only as an assistance to the human decision-maker. A consistency requirement on demand resolution needs to be enforced, viz., that at most one of a pair of conflicting demands (one from each contesting parties) can be of {\small {\sf ACCEPTED}}. This soft approach keeps open the possibility that some demands of one party and some demands of the other party may be {\small {\sf ACCEPTED}}, as discussed above. 
%\end{enumerate}

We point out that our nuanced, assistance-oriented, multi-step  approach to DR (proposed above) is quite different from the supervised classification approaches to decision prediction as reported in the literature~\cite{alcantara2022survey}. There, the focus is on training a judgement prediction model using a large corpus of past court cases, and making a one-step, take-it-or-leave-it prediction of the decision, often without producing any justifications or explanations. In our case, the approach is aligned with the consensual process, and is completely {\em unsupervised}%and {\em isolated}
, in the sense that no corpus of past disputes is used to perform DR in a given new dispute.

DRAssist suggests a first-cut trial resolution of any new incoming dispute, in order to assist the human decision-maker. This approach will also help in reducing time and efforts spent in DR, as well as reduce the biases and subjectivity that might be naturally present in such processes. The system can also refuse to provide a resolution, if it is unable to come up with a ``strong'' justification for any resolution. The resolution recommended by the system need not be binding on the parties, and may undergo a review and refinement by human experts. However, the hope would be that such a system can speed up the resolution time-frames, reduce efforts, provide human-understandable justifications and reduce subjectivity and biases that may affect human adjudicators. More over, with machine learning techniques built into such a system, it will hopefully learn from its mistakes and improve continuously. 

%In this paper, we demonstrate the effectiveness of our approach in resolving disputes in two very different domains: automobile insurance and domain name disputes. The two datasets of disputes in these domains will be released upon acceptance of the paper for publication.  

%\textcolor{red}{TODO: I have removed dispute categories enumeration and examples of these categories in GST domain (Table 1). Add these back into the paper??}

%The paper is organized as follows. 
%\textcolor{red}{TODO: to be completed.}
Our main contributions in this paper are:

%\begin{itemize}
    %\item 
$\bullet$ Datasets of disputes in two different domains -- {\it automobile insurance disputes} and {\it domain name disputes} (Section ~\ref{secDatasets}).
    %\item 
    
$\bullet$ Identification of key structural elements of a dispute and summarizing all the disputes in our datasets in accordance with these elements to produce a {\em structured summary} in a standard format for each dispute (Section~\ref{secStructuredSummary}). Creating suitable {\em ground truth} for each dispute in terms of the winning party, demand-wise decisions, and argument evaluation (Section~\ref{secGroundTruth}). Both the datasets, along with structured summaries and ground truth information would be released upon acceptance of the paper.% for publication.
    %\item 
    
$\bullet$ {\bf DRAssist}: A prototype {\em dispute resolution assistance system} where we explore 3 different LLMs for assistance in resolving the disputes in the two datasets in {\em zero-shot manner} (Section~\ref{secDRSystem}).
%\end{itemize}

\section{Related Work}
\subsection{Disputes in Different Domains}
%As mentioned earlier, disputes occur in many domains. To get a better understanding of the typology, similarities and differences among disputes, and methods to resolve them, we review here work on some domain-specific disputes. 
%We briefly review some past work about disputes in various domains.

\noindent\textbf{E-Commerce Disputes}: %Disagreement between buyers and sellers on large e-commerce platforms are unavoidable. It is important to resolve these disputes in an accurate, fast, and fair manner for maintaining a trustworthy platform. Although, simple cases can be automated, intricate cases are not sufficiently addressed by hard-coded rules. Tsurel et al.~\cite{tsurel2020commerce} took a first step towards automatically assisting human agents in dispute resolution at scale. They constructed a large dataset of disputes from the {\em eBay} online marketplace, and identified several interesting behavioral and linguistic patterns. The data had 937 features which belong to various feature families -- (i) Claim (features related to claim, e.g., {\em first escalating party}), (ii) Transaction (features related to the transaction before the claim, e.g. {\em price of the item}, (iii) Claim seller (claim features related to the seller, e.g., {\em seller tenure days}), (iv) Claim buyer (claim features related to the buyer, e.g., {\em no. of disputer buyer participated in last year}), (v) Seller data (features related to the seller user profile, e.g. {\em city of residence}), (vi) Buyer data (features related to the buyer user profile, e.g., {\em tax status}), (vii) Textual features (derived from the text of conversation between the buyer and the seller, e.g. {\em direct question}, {\em apologizing}). Several classifiers were then trained to predict dispute outcomes with high accuracy. Also, they explored interpretability of features using ELI5 tree explainer for XGBoost as well as LIME.
These disputes occur between buyers and sellers on large e-commerce platforms. Tsurel et al.~\cite{tsurel2020commerce} took a first step towards automatically assisting human agents in dispute resolution at scale. They constructed a large dataset of disputes from the {\em eBay} online marketplace, and identified several interesting behavioral and linguistic patterns. The data had 937 features from various feature families. Some example features are -- {\em first escalating party}, {\em price of the item}, {\em tax status of the buyer}, etc. Several classifiers were then trained to predict dispute outcomes with high accuracy. %Also, they explored interpretability of features using ELI5 tree explainer for XGBoost as well as LIME.

\noindent\textbf{Domain Name Disputes}: %Vihikan et al.~\cite{vihikan2021automatic} introduced a new task which involved predicting the outcome of a process for resolving disputes about legal entitlement to a domain name. These disputes arise between a trademark owner and a domain name registrant pertaining to a generic Top-Level Domain (gTLD) name (e.g., \url{ggroupon.com} which is similar to the trademark ``groupon'' of the company Groupon Inc). In such disputes, the trademark owner (complainant) must show that four conditions are satisfied: (i) they own a trademark, (ii) the domain name in dispute is identical or confusingly similar to that trademark, (iii) the domain name registrant (respondent) has no rights or legitimate interests in the domain name, and (iv) the  domain name registrant registered and has used the domain name in bad faith. The task is introduced as a means to assist the service providers and arbitrators by automatically pre-assessing complaints. The ultimate aim is to identify those complaints whose characteristics are predictive of success. Doing so has the potential to aid in the optimal utilization of the time and expertise of the arbitrator, because such cases should be relatively easier for the arbitrator to decide. They created a corpus of 31013 cases by scraping the website of Word Intellectual Property Organization (WIPO)\footnote{\url{https://www.wipo.int/amc/en/domains/decisionsx/index.html}}. For each case, the description consisted of its factual background and contentions (arguments) by each party. Only 3 majority labels were considered -- TRANSFER, COMPLAINT DENIED, CANCELLATION. They explored multiple classifiers on this dataset based on several models such as BiLSTM, BiGRU, BERT, and LEGAL-BERT.
These disputes arise between a trademark owner and a domain name registrant pertaining to a generic Top-Level Domain (gTLD) name (e.g., \url{ggroupon.com} which is similar to the trademark ``groupon'' of the company Groupon Inc). Vihikan et al.~\cite{vihikan2021automatic} proposed a task to assist the service providers and arbitrators by automatically pre-assessing these disputes. They created a corpus of 31013 cases by scraping the website of Word Intellectual Property Organization (WIPO)\footnote{\url{https://www.wipo.int/amc/en/domains/decisionsx/index.html}}. They explored multiple classifiers on this dataset based on several models such as LEGAL-BERT~\cite{chalkidis2020legal}. As this dataset was not available\footnote{The dataset link is inactive, and the email request to the authors remained unanswered.}, we ourselves scraped the WIPO website to create a new dataset. 
Ferro et al.~\cite{ferro2019scalable} presented an approach to develop a feature-rich corpus of domain name disputes that leverages a small amount of manual annotation and prototypical patterns present in the case documents to automatically extend feature labels to the entire corpus. All the approaches described above use some form of supervision, whereas we avoid any supervision in our proposed technique.
%Ferro et al.~\cite{ferro2019scalable} presented an approach to developing a feature-rich corpus of administrative rulings about domain name disputes. Their approach leverages a small amount of manual annotation and prototypical patterns present in the case documents to automatically extend feature labels to the entire corpus. Here, they use a feature set that makes use of two common elements in legal argumentation: {\em issues} and {\em factors}. They define an ``issue'' to be a formal element of a legal claim corresponding to a term, or ``predicate'' that occurs in some authoritative legal source (e.g., statute). On the other hand, ``factor'' is any aspect for which the decision portion of one or more cases contains findings about to justify conclusions about some {\em issue}. They consider 3 major issues about domain name disputes -- (i) ICS: (domain name is) Identical or Confusingly Similar, (ii) NRLI: No Rights or Legitimate Interests (in domain name), and (iii) BadFaith: domain name is registered and used in bad faith. \textcolor{red}{TODO: dataset details and explanation of Issue-Outcome prediction task.}

\noindent\textbf{Tax Disputes}: Disputes between tax-payers and tax-collecting authorities are common~\cite{But16},~\cite{MGEB24}. These disputes tend to be complex, because of the bewildering variety and complexity of tax laws. For example, Goods and Services Tax (GST) in India~\cite{RM16}, which is similar to the VAT in many countries, faces many types of disputes including: profiteering (not passing the benefits of reduced GST to customers), invoice errors (false documentation, incorrect information), incorrect due GST amount, under-payment of due GST amount, delays in GST payments, incorrect interpretation of clauses in GST Law, disallowed input tax, incorrect qualification of turnover, inconsistencies between GST and annual financial statements, failure to file returns etc. Due to the large volumes of tax disputes, Govt. of India runs specialized DR clinics for direct taxes called {\em Vivad Se Vishwas}\footnote{https://www.livemint.com/money/personal-finance/vivad-se-vishwas-scheme-2024-settle-income-tax-disputes-by-december-31-to-avoid-penalties-11735029235022.html} (literally means {\it from dispute to trust}).

%\noindent\textbf{Insurance Disputes - Sachin}: 

%\noindent\textbf{Banking Disputes - Manoj}: 

\noindent\textbf{Environment Disputes}: %There are often disputes between a national environmental protection agency and a foreign (or local) investor, such as a manufacturing or mining company. If the investor is international, then such disputes are handled by an agency like the Investment Treaty Arbitration. The typology of disputes includes: (i) \textbf{operations of the company} (e.g., waste treatment, garbage, pesticides/chemicals, emissions, land-filling, biodiversity compensation etc.); (ii) \textbf{impact on environment or other entities}; e.g., water/air pollution, displacement or health of people, tourism etc.; (iii) \textbf{applicability of relevant laws}. Behn and Langford~\cite{BL17} analyzes environmental cases to detect evidence of specific biases the arbitration agency may have; e.g., excessive compensations, non-compliance to democratic processes, restricting the environment protection actions etc. 
There are often disputes between a national environmental protection agency and a foreign (or local) investor, such as a manufacturing or mining company. If the investor is international, then such disputes are handled by an agency like the Investment Treaty Arbitration. The typology of disputes includes: (i) {\it operations of the company} (e.g., waste treatment, pesticides/chemicals, emissions, etc.); (ii) {\it impact on environment or other entities} (e.g., water/air pollution); (iii) {\it applicability of relevant laws}. Behn and Langford~\cite{BL17} analyzes environmental cases to detect evidence of specific biases the arbitration agency may have; e.g., excessive compensations, non-compliance to democratic processes, restricting the environment protection actions etc.   

%\noindent\textbf{Landlord-Tenant Disputes for Rent Reduction}: Westermann et al.~\cite{westermann2019using} identified 44 factors that occur in disputes where the tenant seeks a remedy due to problems with the rented apartment, such as the existence of bedbugs, high noise levels or problems with insulation. They analyzed the correlation between these factors found in a case and the final decision of the judge (the amount of reduction awarded). 

%\noindent\textbf{Similarities and Differences in Disputes across Domains}: TBD

%\section{Related Work}

%We review some of the related research.

\subsection{AI-based Dispute Resolution}

% \cite{guan2024predicting} focuses on predicting a path (sequential methods) to resolve labour dispute cases. The methods include 1) mediation in arbitration, 2) arbitration awards, 3) first-instance mediation, 4) first-instance judgments, and 5) second-instance judgments. These ordered sequence of methods forms a path such that it will start from method 1 and could halt at any other method say k (where $k \in [1,5]$) to resolve any dispute. Their can be total 5 possible path of lengths L (where $L \in [1,5]$) of labor dispute resolution, these path length's are considered as labels for a legal dispute case, thus translating this task to multi-class classification. Predicting the optimal critical path of dispute resolution can save lots of time and resources of dispute resolution institutions. They applied a machine learning approach named LDMLSV, based on SHAP and a soft voting strategy, to predict the critical path of labor dispute resolution where they tailored the features from the dispute cases and employed 10 different machine learning models, where they selected three models whose accuracies are greater than 0.85. In addition, they also ensembled these models called as soft voting model which aggregates the outputs of these models. Furthermore, they selected crucial features influencing these models using Shapley values and retrained the soft voting classifier which achieves an accuracy of 0.90.

Guan et al.~\cite{guan2024predicting} focuses on predicting a sequence of methods to resolve labor disputes, including mediation in arbitration, arbitration awards, first-instance mediation, first-instance judgments, and second-instance judgments. The path starts at method 1 and can halt at any method (where $k \in [1,5]$), forming 5 possible paths of lengths (where $L \in [1,5]$), which are used as labels for multi-class classification. Predicting the optimal path can save time and resources for dispute resolution institutions. They applied the LDMLSV machine learning approach, based on SHAP and soft voting, to predict the critical path. After tailoring features from dispute cases and employing 10 models, they selected three with accuracies over 0.85. The final soft voting model, retrained with Shapley values, achieved an accuracy of 0.90.

Yang et al.~\cite{yang2019detecting} go beyond the negative sentiments and dissatisfaction expressed by customers of an e-commerce company in their chats with a customer service chatbot and tackle the problem of predicting the customer's {\em intent} from their dialog sequences about whether they plan to escalate the grievance into a formal complaint with government authorities. They use a single-layer RNN (with attention mechanism), augmented with additional textual features (such as \#emojis, \#question marks, \#exclamation marks, \#ellipsis, sentiment score, TF-IDF scores of the dialog based only on the terms from a domain dictionary etc.). Only 21K dialogues (out of a million) were escalated to formal complaints. They used a 300-dim embedding, and used cross-entropy loss and Adam optimizer to train the model. 
%Since most escalated complaints are filed within a week of the last contact with the customer, they evaluate the recall over a 7-day window and show that their model performs better than reasonable baselines like Logistic Regression and a tree-based classifier. 
%The work is not about automated dispute resolution, nor does it offer detailed insights into why some disputes are escalated and some are not.
The work is not about dispute resolution, nor does it offer detailed insights into why some disputes are escalated and some are not.
%Although not really in the scope of this paper, other domains contain interesting types of ``disputes'',  prominent among them being political, social and economic disputes on a large-scale. \cite{zaczynska2024diplomats} provides a database of disagreements or critiques in the diplomats' speeches in UN security council, which often refer to some political conflicts. They annotate 87 speeches about Crimea annexation and Ukraine war with categories of disagreements, such as {\sf Direct Negative Evaluation}, (a direct verbal attack on the political position of another country), {\sf Indirect Negative Evaluation}, {\sf Challenge} (accusing another country of not telling the truth) and {\sf Correction} (correcting  a false statement made by another country). Other parts of the dispute, such as Source and Target Country, are also annotated.
%\textcolor{red}{TODO: review this paper? ~\cite{zeleznikow2021using}}

\vspace{-2mm}
\subsection{Predicting Judicial Decisions}
In this paper, we have proposed a nuanced, assistance-oriented multi-step DR process by taking into account the positions of both sides, which focuses on resolving demands of both parties and suggesting a stronger party. We review a few research works related to judgement prediction legal cases, which can be considered a {\em hard} approach to DR; see the survey~\cite{alcantara2022survey}. 
Normally each ``simple'' case in the US Supreme Court is heard by a bench consisting of 9 judges. Guimera and Sales-Pardo~\cite{GS11} attempt to predict a judge's vote, given the votes of other 8 judges in the current ($n^{th}$) case and votes of all 9 judges in last $n-1$ cases, ignoring the actual contents of the case. Based on the observation that in real courts the judges' votes are correlated, they divide cases and judges into all possible {\em blocks}, and provide a function which averages the observed probabilities (computed from the past cases) in which a judge from block $a$ voted in favour of the petitioner in the case in block $b$. This block model algorithm correctly predicts 83\% of the individual judges’ decisions in the dataset of US Supreme Court cases. However, in our scenario, the DR is expected to be done by a single arbitrator, not by a bench. 

Medvedeva et al.~\cite{MVW20} use judgements in Human Rights violation cases of the European Court of Human Rights. Their goal is to predict which (if any) of the 9 articles of the European Convention on Human Rights are found to be violated in a particular case. The articles refer to categories of violations such as {\sf Torture, Slavery and Forced Labour, Fair Trial, Freedom of Expression} etc. They use supervised SVM with n-gram features, and report an average prediction accuracy of 75\%. They also report a common structure of the cases, which is similar to the structure of disputes that we have proposed in this paper.  

LLMs play a crucial role in our proposed DR process. Shui et al.~\cite{shui2023comprehensive} evaluates the effectiveness of LLMs like GPT-4, Vicuna, ChatGLM, and BLOOMZ in predicting legal judgments based on complex case data. The focus is on ability of these models to handle tasks requiring domain-specific legal reasoning and compare the performance in different settings like standalone LLM and with assistance from an Information Retrieval (IR) system which supplies relevant cases or outcome suggestions (``label candidates'') to questions that can be used in the prompts to improve the contextual understanding of the LLM. Results show that the IR assisted LLM setting gives better results and predict judgements more accurately than standalone LLMs. However they also present a paradoxical scenario where the IR system on its own can surpass the performance of the LLM+IR system where the LLM is taking on a leading role and IR is supporting it. The results are mixed where GPT-4 and ChatGPT show remarkable proficiency about the domain knowledge, ChatGLM a smaller LLM shows greater robustness whereas BLOOMZ showcased better zero shot ability. 

%\noindent\textbf{Some very generic papers}: ~\cite{jacobsen2023smart,barton2020artificial,candeias2023artificial}

\section{Datasets}\label{secDatasets}
We explored two datasets %\footnote{Both the datasets would be made available upon publication.} 
for our experiments -- auto-insurance disputes and domain name disputes. %In this section, we discuss these two datasets in detail.

%\subsection{Auto-insurance Disputes ($D_{AI}$)}
\noindent\textbf{Auto-insurance Disputes (\bm{$D_{AI}$})}: 
%We collected 129 court cases (dataset $D_1$) in the automobile insurance domain from the Supreme Court of India\footnote{http://liiofindia.org/in/cases/cen/INSC/}. These court cases are very similar in structure to disputes between an insurance company and a vehicle owner. The summary statistics of this corpus is as follows:\\*
We collected 1117 insurance disputes from the National Consumer Disputes Redressal Commission (NCDRC), India. These disputes are downloaded from the ConfoNet project website\footnote{\url{https://cms.nic.in/ncdrcusersWeb/search.do?method=loadSearchPub} (accessed on 30-APR-2024)} by selecting the sector as ``Insurance'' for the years 2022, 2023, and 2024. These disputes are between an insurance company and an insured party. We kept only automobile insurance disputes by removing disputes belonging to other type of insurance (e.g., health). We also removed the disputes where there was no clear winning party (15\%) so that the focus remains on the disputes where the human judgement was clearly decisive. Finally, we ended up with the dataset of \textbf{104} disputes referred to as $D_{AI}$. %For this dataset, the applicable legal statute usually is the Insurance Act, 1938\footnote{https://www.indiacode.nic.in/bitstream/123456789/2304/1/a1938-04.pdf}.

%\subsection{Domain Name Disputes ($D_{DN}$)}
\noindent\textbf{Domain Name Disputes (\bm{$D_{DN}$})}:
Following Vihikan et al.~\cite{vihikan2021automatic}, we scraped all the 3681 disputes in English that were resolved in 2021 by World Intellectual Property Organization (WIPO) from their webpage\footnote{\url{https://www.wipo.int/amc/en/domains/decisionsx/index.html}}. These disputes are between a complainant (who holds a copyright to a term used in a domain name) and a respondent (who is the registrant of the domain name). Very often the respondent does not formally respond to the complaint, resulting in the domain name getting transferred to the complainant. Such disputes are not very interesting because there are no formal arguments from the respondent's side. Hence, we retained only \textbf{351} of these disputes where the respondent has formally submitted his/her arguments. We refer to this dataset of 351 domain name disputes as $D_{DN}$.

\begin{figure}[]
\centering
\includegraphics[width=0.95\columnwidth,height=0.55\columnwidth]{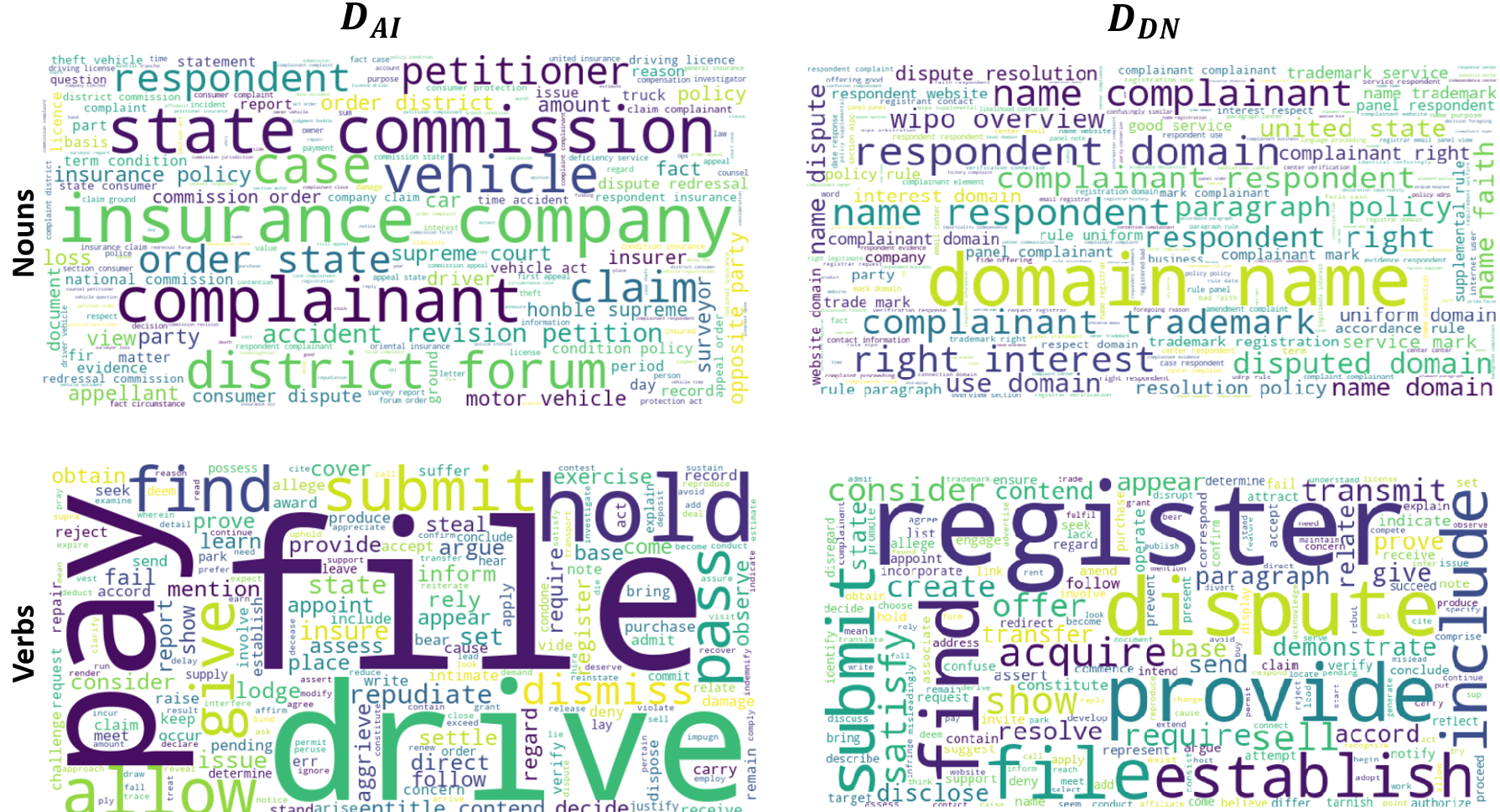}
\caption{Noun and verb word clouds for the datasets $D_{AI}$ (auto-insurance disputes) and $D_{DN}$ (domain name disputes)} \label{fig_wc}
\end{figure}

\begin{figure}[]
\centering
\includegraphics[width=0.9\columnwidth,height=0.41\columnwidth]{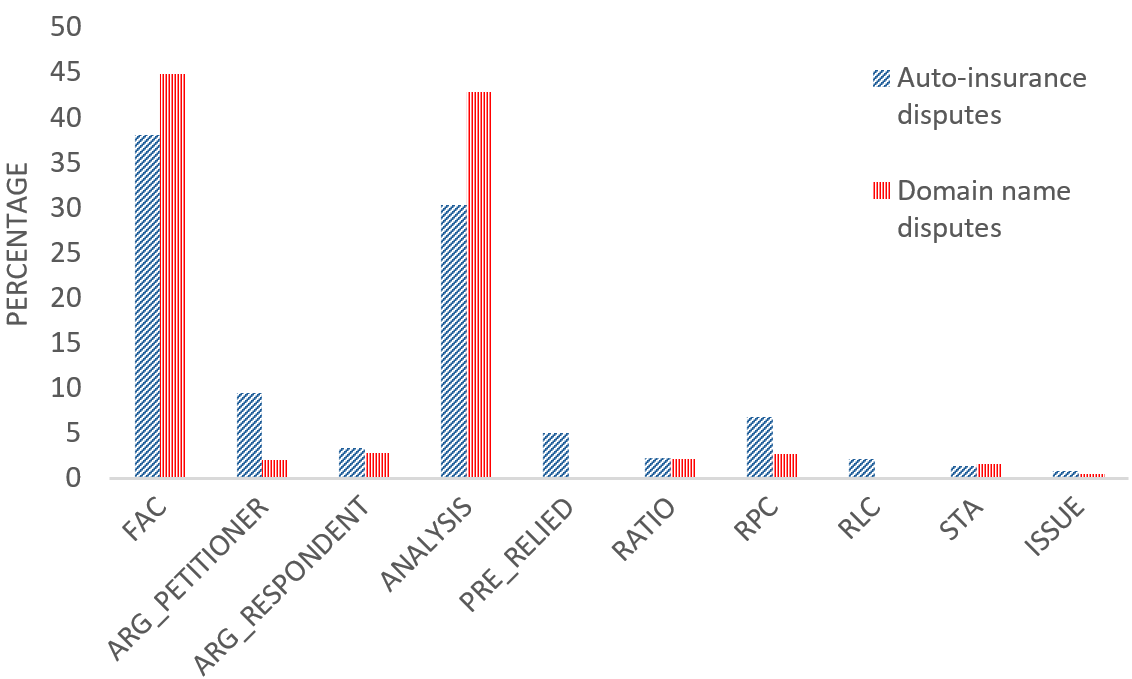}
%\caption{Relative prevalence of rhetorical roles in the datasets $D_{AI}$ (auto-insurance disputes) and $D_{DN}$ (domain name disputes)} \label{fig_rr}
\caption{Relative prevalence of rhetorical roles in the datasets $D_{AI}$ and $D_{DN}$} \label{fig_rr}
\end{figure}

%\subsection{Dataset statistics}
\noindent\textbf{Dataset statistics}:
\begin{table}[]\small
    \centering
    \caption{Dataset statistics}
    \label{tabDatasetStats}
    \begin{tabular}{llrr}
    \hline
    & {\bf Statistic} & \bm{$D_{AI}$} & \bm{$D_{DN}$} \\
    \hline
    No. of disputes (documents) & - & 104 & 351 \\
    \hline
    \multirow{5}{*}{\parbox{2.5cm}{No. of sentences per document}} & Mean & 78.9 & 121.2 \\
    & Std. Deviation & 33.8 & 46.7 \\
    & First Quartile (Q1) & 56.0 & 88.5 \\
    & Median (Q2) & 75.0 & 113.0 \\
    & Third Quartile (Q3) & 96.3 & 140.5 \\
    \hline
    \multirow{5}{*}{\parbox{2.5cm}{No. of words per document}} & Mean & 2343.1 & 3448.8 \\
    & Std. Deviation & 982.7 & 1514.2 \\
    & First Quartile (Q1) & 1610.8 & 2435.0 \\
    & Median (Q2) & 2164.0 & 3196.0 \\
    & Third Quartile (Q3) & 2802.8 & 4082.5 \\
    \hline
    \end{tabular}
\end{table}
Table~\ref{tabDatasetStats} shows the basic dataset statistics for $D_{AI}$ and $D_{DN}$. 
% The summary statistics of this corpus is as follows:\\*
% %For \#sentences per document: average:81.8 STDEV:39.1 first quartile:57 median:79 third quartile:95\\*
% %For \#words per document: average:2475.7 STDEV:1174.5 first  quartile:1635 median:2221 third quartile:2792\\*
% For \#sentences per document: average:78.1 STDEV:32.5 first quartile:56 median:77 third quartile:93\\*
% For \#words per document: average:2352 STDEV:950.8 first  quartile:1634 median:2199 third quartile:2792\\*
Figure~\ref{fig_wc} shows word clouds of prominent nouns and verbs occurring in these datasets. To get an understanding about what types of sentences occur in disputes, we used the Python package {\tt opennyai}\footnote{https://pypi.org/project/opennyai/} based on \cite{KTAK22} to assign a {\em rhetorical role} for each sentence in each dispute in $D_{AI}$ and $D_{DN}$ (Figure~\ref{fig_rr}). It can be observed that majority of the sentences in both the datasets belong to the following rhetorical roles -- FAC (facts), ANALYSIS, and ARG\_PETITIONER/ARG\_RESPONDENT (arguments by both the parties). Also, unlike $D_{AI}$, $D_{DN}$ rarely contains any sentence belonging to PRE\_RELIED (prior cases) and RLC (rulings by lower courts/forums) roles. 

\section{DRAssist: A Prototype DR Assistance System}\label{secDRSystem}
In this section, we describe two main components of DRAssist -- structured summarization and DR assistance.

\subsection{Structured Summarization}\label{secStructuredSummary}
Though the disputes in our datasets roughly follow the same outline, they are quite unstructured. We identified a set of important commonly occurring structural elements of any dispute and summarized each dispute in both the datasets such that all the disputes follow the same {\em standardized structure}. Following are these structural elements for $D_{AI}$:
\begin{enumerate}
    \item {\em Facts agreed by both parties}
    \item {\em Aspects on which the parties disagree}
    \item {\em Demands of the insurance company}
    \item {\em Demands of the insured party}
    \item {\em Arguments of the insurance company}
    \item {\em Arguments of the insured party}
    \item {\em Relevant prior cases referred with short summary} %(if mentioned)
    \item {\em Relevant statutes or policy terms and conditions referred with short summary} %(if mentioned)
    \item {\em Decision by District Commission} %(if mentioned)
    \item {\em Decision by State Commission} %(if mentioned)
    \item {\em Final decision by the National Commission with respect to each demand of both the parties}
    \item {\em Justification / rationale for the final decision}
    \item {\em Winning party}
\end{enumerate}
Except for the points 9 and 10 above (decision by district/state commission) that are specific to $D_{AI}$, other points are quite general and applicable for other types of disputes like $D_{DN}$. The only other differences in the structure of $D_{DN}$ are -- (i) the names of contesting parties are changed accordingly ({\em insurance company} $\leftrightarrow$ {\em complainant} and {\em insured party} $\leftrightarrow$ {\em respondent}), and (ii) the structural element of prior cases is dropped as it is not observed commonly in the domain name disputes. To convert the original disputes to a structured summary following the above structure, we used 3 different LLMs -- 2 open-source medium sized LLMs (Mistral\footnote{\url{https://huggingface.co/mistralai/Mistral-7B-Instruct-v0.3}}, Llama\footnote{\url{https://huggingface.co/meta-llama/Meta-Llama-3-8B-Instruct}}) and 1 closed-source LLM (GPT-4o-mini). The LLM is prompted with the entire unstructured dispute details along with the above structural elements to produce a structured summary of the given dispute. On qualitative evaluation of these structured summaries, we observed that any summary prepared by an individual LLM may miss some important piece of information or it may hallucinate in some rare cases. In order to alleviate these issues, we generate a super summary by merging the summaries generated by the 3 LLMs. This super summary is also generated by prompting an LLM with the 3 candidate summaries (for a specific structural element at a time), asking it to merge these summaries so that the merged summary is consistent with the majority of the individual summaries. We used GPT-4o-mini for generating this super summary in a step-by-step manner, by merging each structural element (except ``Winning party'') at a time in each step. For deciding the winning party in the super summary, we consider the majority vote between the winning parties identified by the 3 individual summaries. Finally, out of the 104 disputes, in 35 disputes (33.7\%) ``insurance company'' is the winner and in 69 disputes (66.3\%) ``insured party'' is the winner. See Table~\ref{tabStructuredSummary} for the structured summary of a sample dispute. On the other hand, as the final result for the disputes in $D_{DN}$ are mentioned in more consistent language, we used a simple regular expression to determine the winning party. Out of 351 disputes in $D_{DN}$, in 234 disputes (66.7\%), ``complainant'' (copyright owner) is the winner and in 117 disputes (33.3\%), ``respondent'' (registrant of the domain name) is the winner.
\begin{table}[t]\small
    \centering
    %\caption{Evaluating structured summaries (R1-F1: F1 for ROUGE-1, RL-F1: F1 for ROUGE-L, BS-F1: F1 for BERTScore)}
    \caption{Evaluating structured summaries (F1-scores for ROUGE-1, ROUGE-L, and BERTScore)}
    \label{tabEvalStructuredSummary}
    \begin{tabular}{cp{3.5cm}ccc}
    \hline
    \textbf{Dataset} & \textbf{Structural Elements} & \textbf{R1-F1} & \textbf{RL-F1} & \textbf{BS-F1} \\
    \hline
    \multirow{4}{*}{$D_{AI}$} & Facts \& Disagreement Aspects & 0.35 & 0.22 & 0.63 \\
    \cline{2-5}
    & Arguments & 0.40 & 0.22 & 0.62 \\
    \cline{2-5}
    & Prior Cases & 0.42 & 0.27 & 0.64 \\
    \cline{2-5}
    & Statutes & 0.35 & 0.22 & 0.57 \\
    \hline
    \multirow{4}{*}{$D_{DN}$} & Facts \& Disagreement Aspects & 0.30 & 0.19 & 0.62 \\
    \cline{2-5}
    & Arguments & 0.41 & 0.24 & 0.64 \\
    \cline{2-5}
    & Statutes & 0.30 & 0.18 & 0.51 \\
    \hline
    \end{tabular}
\end{table}

%In order 
To evaluate the quality of the structured summaries, we could not employ the standard summarization evaluation because of unavailability of gold-standard reference summaries. Hence, we decided to use an approximate estimate of the quality with the help of the rhetorical roles~\cite{KTAK22} identified for individual sentences in the original disputes text. Sentences belonging to certain types of rhetorical roles were considered as {\em reference summary} for a specific type of structural element (or group of elements). For example, the sentences in original dispute text with the rhetorical roles as FAC and ISSUE were considered as reference for our structural elements of agreed facts and disagreement aspects. Similarly, other such mappings considered between rhetorical role labels and our structural elements are as follows: \{{\small ARG\_PETITIONER}, {\small ARG\_RESPONDENT}\} $\leftrightarrow$ arguments by both the parties, \{{\small PRE\_RELIED}, {\small PRE\_NOT\_RELIED}\} $\leftrightarrow$ prior cases, \{STA\} $\leftrightarrow$ relevant statutes. Table~\ref{tabEvalStructuredSummary} shows the evaluation results for multiple structured summary elements where each is compared with corresponding rhetorical role references. Please note that this is only an approximate estimate of summarization quality because the rhetorical role labels (obtained using {\tt opennyai}) themselves may not be perfect.
%TODO: Add similar details for the domain name disputes dataset
%TODO: Ground truth creation for demand-wise as well as argument-wise predictions.
%insurance company  35
%insured party  69
%partial    21

\subsection{Using LLMs for DR Assistance}
Once any dispute is represented in a standardized structured form as described above, DRAssist uses LLMs to produce the desired resolution output. 
%We explore a prototype DR assistance system for assisting human decision-makers for resolution of disputes. 
Input to the resolution component of DRAssist is the description of a dispute represented in this standardized structured form.  %(excluding structural elements corresponding to intermediate or final decisions). 
Please note that we do not include the structural elements corresponding to intermediate or final decisions in this input to an LLM to avoid influencing its judgement. 
For any existing dispute, if the dispute description is available in textual format, a structured summary is generated first whereas for any new dispute, the information about it can be entered by the user in a template following the same structure. 
%Output of the DR system would be a judgement on which of the two contesting parties (insurance company or the insured party in this case) is more likely to be the ultimate winner, along with suitable justification for the same.
Output of the DR system for a given dispute is generated at multiple levels as follows:
%\begin{enumerate}
    %\item 
    
\noindent $\bullet$ {\bf Overall stronger party}: Between the two contesting parties in the given dispute, predicting the overall stronger party along with a suitable justification. Here, the {\em stronger party} refers to the party which is more likely to win given the current facts, arguments, etc.

    %\item 
\noindent $\bullet$ {\bf Demand-wise decisions}: For each demand of each contesting party, predicting whether the demand should be {\em accepted} or {\em rejected}, along with suitable justification.

    %\item 
\noindent $\bullet$ {\bf Argument-wise evaluation}: As part of intermediate reasoning, each argument of each contesting party is evaluated as {\em strong} or {\em weak}, along with suitable justification.
%\end{enumerate}

In this paper, our goal is to evaluate the capabilities of some representative recent LLMs for DR assistance in zero-shot manner. Here, we expect the LLMs to use their inherent world knowledge obtained during their pre-training for analysing different aspects of a given dispute and generating the expected output on multiple levels as described above. The LLM is expected to analyze the given dispute that is fed to it in terms of its multiple structural elements (such as agreed facts, disagreement aspects, arguments, prior cases cited, relevant statutes or terms and conditions referred, etc.), compare and evaluate these elements for their relative strengths, and then generate the expected output in a zero-shot manner. We explored the following three prompting strategies.

\noindent \bm{$S_1$} \textbf{(Direct prompting)}: Here, the expected output is only the overall stronger party. The prompt describes any dispute in terms of the following structural elements - agreed facts, disagreement aspects, arguments, prior cases cited, relevant statutes or terms and conditions referred. The only instruction in the prompt is to predict the overall stronger party along with some justification.

\noindent \bm{$S_2$} \textbf{(Prompting for fine-grained demand-wise decisions)}: Here, the expected output is not only the overall stronger party but also more fine-grained predictions containing demand-wise decisions. Each demand of each party is analyzed to predict whether it is acceptable or not. The prompt describes the dispute using the same structural elements as above, along with the list of demands of each party. The prompt contains two instructions -- (i) for predicting the overall stronger party and (ii) for predicting acceptability of each demand of each party.

\noindent \bm{$S_3$} \textbf{(Chain-of-Thought)}: Here, the expected output contains the overall stronger party as well as the demand-wise decisions similar to $S_2$. In addition, the output is also supposed to contain evaluation of each argument. This prompting strategy is motivated from the zero-shot Chain-of-Thought (CoT) prompting strategy proposed by Kojima et al.~\cite{kojima2022large} in the sense that the LLM is first asked to build a reasoning path through evaluation of arguments rather than directly predicting the stronger party. The prompt also clearly explains {\em why a certain argument should be evaluated as strong or weak} (see Table~\ref{tabResolutionPrompt}). The prompt describes the dispute using the same structural elements as above. It now contains {\em three} instructions -- (i) for evaluating whether each argument of each contesting party is {\em strong} or {\em weak}, (ii) for predicting the overall stronger party based on the evaluation of arguments of both the parties, and (iii) for predicting acceptability of each demand of each party depending on strength or weakness of the relevant arguments.

% Table~\ref{tabResolutionPrompt} shows the prompt used for resolution of auto-insurance disputes for the prompting strategy $S_3$ (Chain-of-Thought). The prompt used in case of domain name disputes is same except the names of the contesting parties ({\em complainant} in place of {\em insurance company} and {\em respondent} in place of {\em insured party}). The prompts used for the strategies $S_1$ and $S_2$ differ with respect to the instructions and response formats. For $S_2$, only the last two instructions are used whereas for $S_1$ only the second instruction is used (as per Table~\ref{tabResolutionPrompt}). Also, the instructions for $S_1$ and $S_2$ do not mention anything about arguments or their strengths. For $S_1$, the dispute description is also different in the sense that it does not include demands of both the parties. The response formats for $S_1$ and $S_2$ are also different in accordance with their respective expected outputs. We only mention the $S_3$ prompt in the paper and do not explicitly show the prompts used for $S_1$ and $S_2$ due to space constraint and also because $S_3$ prompt subsumes them in some way. 
% Table~\ref{tabResolutionOutput} shows the resolution output generated by an LLM using $S_3$.
Table~\ref{tabResolutionPrompt} shows the prompt used for resolution of auto-insurance disputes for the prompting strategy $S_3$ (Chain-of-Thought). The prompt used in case of domain name disputes is same except the names of the contesting parties ({\em complainant} in place of {\em insurance company} and {\em respondent} in place of {\em insured party}). We only mention the $S_3$ prompt in the paper and do not explicitly show the prompts used for $S_1$ and $S_2$ due to space constraint and also because $S_3$ prompt subsumes them in some way. The $S_1$ and $S_2$ prompts differ from the $S_3$ prompt with respect to the instructions and response formats. For $S_2$, only the last two instructions are used whereas for $S_1$ only the second instruction is used (as per Table~\ref{tabResolutionPrompt}). Also, the instructions for $S_1$ and $S_2$ do not mention anything about arguments or their strengths. For $S_1$, the dispute description is also different in the sense that it does not include demands of both the parties. The response formats for $S_1$ and $S_2$ are also different in accordance with their respective expected outputs. %
Table~\ref{tabResolutionOutput} shows the resolution output generated by an LLM using $S_3$.

\begin{table}[t]\footnotesize
    \centering
    \caption{LLM prompt used for the prompting strategy $S_3$ in case of auto-insurance disputes.}
    \label{tabResolutionPrompt}
    \begin{tabular}{p{0.97\columnwidth}}
    \hline
    Consider the following dispute between an insurance company and an insured party.\newline
\textbf{Dispute description}: 

\textbf{Facts agreed by both parties}: {\color{blue}\it \{facts\}}%\newline

\textbf{Aspects on which the parties disagree}: {\color{blue}\it \{disagreement\_aspects\}}%\newline

\textbf{Arguments of the insurance company}: {\color{blue}\it \{arguments\_of\_insurance\_company\}}%\newline

\textbf{Arguments of the insured party}: {\color{blue}\it \{arguments\_of\_insured\_party\}}%\newline

\textbf{Relevant prior cases}: {\color{blue}\it \{prior\_cases\}}%\newline

\textbf{Relevant statutes or policy terms and conditions}: {\color{blue}\it \{statutes\}}%\newline

\textbf{Demands of the insurance company}: {\color{blue}\it \{demands\_of\_insurance\_company\}}%\newline

\textbf{Demands of the insured party}: {\color{blue}\it \{demands\_of\_insured\_party\}}\newline

\#\#\# \textbf{Instructions}:\newline
1. Evaluate each argument by both the contesting parties either as STRONG or WEAK. A STRONG argument is supported by verifiable facts and credible evidence, is logically coherent and internally consistent. In addition, a STRONG argument is supported by either some prior case or some statute or policy terms and conditions. Also, a STRONG argument is legally more sound, provides clear interpretation of reality, and likely to influence the final decision than a WEAK argument. Re-write each argument. Label it as either STRONG or WEAK based on your analysis of the above dispute details, along with short justification.\newline
2. Depending on overall strength of the arguments, identify the Overall Stronger Party (insurance company or insured party). Include a short justification considering the fairness and logic of the dispute resolution.\newline
3. Depending on whether the relevant arguments of each party are STRONG or WEAK, evaluate each demand of each party as either ACCEPTED or REJECTED using the agreed facts, disagreement aspects, arguments, prior cases and statutes or policy terms and conditions. Re-write each demand. Label it as either ACCEPTED or REJECTED based on your evaluation, along with short justification. Ensure justifications are logical and clear. For opposing or conflicting demands, ensure they do not receive the same label (e.g., if one party’s demand is ACCEPTED, the opposing or conflicting demand of the other party should be REJECTED).\newline

Follow the following response format and do not generate any additional output.%\newline

\#\#\# \textbf{Response Format}:\newline
\textbf{Arguments of the insurance company}:\newline
<Argument>: <STRONG or WEAK> : <justification>%\newline

\textbf{Arguments of the insured party}:\newline
<Argument>: <STRONG or WEAK> : <justification>%\newline

\textbf{Overall Stronger Party}: <insurance company or insured party>: <Justification>%\newline

\textbf{Demands of the insurance company}:\newline
<Demand>: <ACCEPTED or REJECTED> : <Justification>%\newline

\textbf{Demands of the insured party}:\newline
<Demand>: <ACCEPTED or REJECTED> : <Justification>\\
    \hline
    \end{tabular}
\end{table}

\begin{table}[]\footnotesize
    \centering
    \caption{Resolution output generated by GPT-4o-mini for the example case in Table~\ref{tabStructuredSummary} using prompting strategy $S_3$ (CoT)}
    \label{tabResolutionOutput}
    \begin{tabular}{p{0.95\columnwidth}}
    \hline
\textbf{Arguments of the insurance company:}\newline
1. The insurance company asserts that the vehicle in question was stolen prior to the issuance of the insurance policy, specifically on 18.12.2002: \textbf{WEAK} : This argument lacks verifiable evidence to support the claim of the theft date and relies solely on the company's assertion without corroborating evidence.%\newline

2. The insured party allegedly misrepresented the vehicle during the inspection process by presenting a different vehicle, leading to the fraudulent acquisition of the insurance policy on 19.12.2002: \textbf{WEAK} : There is no evidence provided to substantiate the claim of misrepresentation, and the insurance company has acknowledged negligence in the inspection process.%\newline

3. The insurance company has acknowledged negligence on the part of its functionary, who failed to conduct a proper physical inspection of the vehicle, and has taken departmental action against this individual for their oversight: \textbf{STRONG} : This admission of negligence indicates a failure in the insurance company's process, which could undermine their position regarding the validity of the policy.%\newline

4. Based on these grounds, the insurance company contests the validity of the insurance policy, claiming it was obtained under fraudulent circumstances: \textbf{WEAK} : The argument is not supported by concrete evidence of fraud, especially given the acknowledgment of negligence in the inspection process.\newline

\textbf{Arguments of the insured party:}\newline
1. The insured party purchased the vehicle on 29.11.2002, with the assurance from the dealer that the insurance policy would be obtained on his behalf: \textbf{STRONG} : This statement is supported by the timeline of events and indicates the insured party's reliance on the dealer's assurance.%\newline

2. Due to delays from the dealer in securing the insurance policy, the insured party took the initiative to obtain the policy himself on 19.12.2002: \textbf{STRONG} : This argument is logical and demonstrates the insured party's proactive approach to securing coverage.%\newline

3. The insured party asserts that the vehicle was shown to the functionary of the insurance company at the time the policy was obtained: \textbf{STRONG} : This claim is reasonable and aligns with the insured party's actions to ensure proper coverage.%\newline

4. The vehicle was stolen from Bodhgaya on 23.12.2002, after the insurance policy was in effect: \textbf{STRONG} : This argument is supported by the timeline and the insured party's claim, which is further backed by the police report.%\newline

5. The insured party claims that the police investigation into the theft confirmed the incident, and the police report has been accepted by the competent court, supporting the validity of the insured party's claims regarding the theft: \textbf{STRONG} : This argument is robust as it references an official investigation and court acceptance, providing credible evidence for the insured party's claims.\newline

\textbf{Overall Stronger Party: insured party}: The insured party presents a more coherent and evidence-backed case, particularly with the support of the police investigation and the timeline of events, while the insurance company's arguments are largely unsupported and rely on assertions without substantial evidence.\newline

\textbf{Demands of the insurance company:}\newline
1. Denial of the insurance claim: \textbf{REJECTED} : The denial is not supported by strong evidence, especially given the insured party's valid claims and the acknowledgment of negligence by the insurance company.%\newline

2. Assertion that the policy was fraudulently obtained: \textbf{REJECTED} : There is insufficient evidence to support the claim of fraud, particularly in light of the company's own admission of negligence.%\newline

3. Repudiation of the claim based on the fraudulent nature of the policy acquisition: \textbf{REJECTED} : The lack of evidence for fraud and the acknowledgment of negligence undermine this demand.\newline

\textbf{Demands of the insured party:}\newline
1. Payment of the insured amount for the stolen vehicle: \textbf{ACCEPTED} : The insured party has provided a strong case supported by the timeline, police report, and the validity of the insurance policy.%\newline

2. Interest on the insured amount at the rate of 10\% per annum: \textbf{ACCEPTED} : Given the acceptance of the claim for the insured amount, it is reasonable to award interest on the amount due, as it compensates for the delay in payment.\\
    \hline
    \end{tabular}
\end{table}

\vspace{-4mm}
\section{Experiments}
\begin{table*}[]\small
    \centering
    \caption{Evaluation results for the datasets $D_{AI}$ (auto-insurance disputes) and $D_{DN}$ (domain name disputes). The three prompting strategies are - $S_1$: Prediction of stronger party (only), $S_2$: Prediction of stronger party and demand-wise decisions, $S_3$: Evaluating of Arguments followed by the prediction of stronger party and demand-wise decisions)}
    %\label{tabResultsVehicleInsurance}
    \label{tabResults}
    \begin{tabular}{ccccccccc}
    \hline
    \multirow{2}{*}{\bf Dataset} & \multirow{2}{*}{\bf Technique} & \multirow{2}{*}{\bf Model} & \multicolumn{2}{c}{\bf Stronger Party Prediction} & \multicolumn{2}{c}{\bf Demand-wise Decision} & \multicolumn{2}{c}{\bf Argument-wise Evaluation} \\
    \cline{4-9}
     &  &  & {\bf Accuracy} & {\bf Macro-F1} & {\bf Accuracy} & {\bf Macro-F1} & {\bf Accuracy} & {\bf Macro-F1} \\
    \hline
    \multirow{14}{*}{$D_{AI}$} & \multirow{2}{*}{Baselines} & Majority Label & 0.66 & 0.40 & 0.60 & 0.38 & 0.56 & 0.36 \\
     &  & Random Label & 0.59 & 0.53 & 0.49 & 0.49 & 0.52 & 0.52 \\
    \cline{2-9}
     & \multirow{4}{*}{$S_1$: Direct prompting} & Mistral-7B-Instruct & 0.44 & 0.42 & - & - & - & - \\
     &  & GPT-4o-mini & 0.69 & 0.68 & - & - & - & - \\
     &  & Llama-3-8B-Instruct & 0.71 & 0.70 & - & - & - & - \\
     %\rowcolor{Gray}
     &  & \cellcolor{Gray}Ensemble (above 3) & \cellcolor{Gray}0.67 & \cellcolor{Gray}0.67 & \cellcolor{Gray}- & \cellcolor{Gray}- & \cellcolor{Gray}- & \cellcolor{Gray}- \\
    \cline{2-9}
     & \multirow{4}{*}{$S_2$: Fine-grained} & Mistral-7B-Instruct & 0.45 & 0.42 & 0.54 & 0.50 & - & - \\
     &  & GPT-4o-mini & 0.78 & 0.75 & 0.60 & 0.62 & - & - \\
     &  & Llama-3-8B-Instruct & 0.69 & 0.69 & 0.62 & 0.62 & - & - \\
     %\rowcolor{Gray}
     &  & \cellcolor{Gray}Ensemble (above 3) & \cellcolor{Gray}0.70 & \cellcolor{Gray}0.70 & \cellcolor{Gray}0.63 & \cellcolor{Gray}0.63 & \cellcolor{Gray}- & \cellcolor{Gray}- \\
    \cline{2-9}
     &  \multirow{4}{*}{$S_3$: Chain-Of-Thought} & Mistral-7B-Instruct & 0.77 & 0.76 & \textbf{0.63} & \textbf{0.64} & 0.57 & 0.58 \\
     &  & GPT-4o-mini & 0.70 & 0.70 & 0.58 & 0.60 & 0.58 & 0.59 \\
     &  & Llama-3-8B-Instruct & 0.74 & 0.67 & 0.55 & 0.57 & 0.55 & 0.59 \\
     %\rowcolor{Gray}
     &  & \cellcolor{Gray}Ensemble (above 3) & \cellcolor{Gray}\textbf{0.80} & \cellcolor{Gray}\textbf{0.78} & \cellcolor{Gray}0.61 & \cellcolor{Gray}0.62 & \cellcolor{Gray}\textbf{0.59} & \cellcolor{Gray}\textbf{0.60} \\
    \hline
    \hline
    \multirow{14}{*}{$D_{DN}$} & \multirow{2}{*}{Baselines} & Majority Label & 0.67 & 0.40 & 0.58 & 0.37 & 0.54 & 0.35 \\
     &  & Random Label & 0.51 & 0.46 & 0.51 & 0.51 & 0.51 & 0.51 \\
    \cline{2-9}
     & \multirow{4}{*}{$S_1$: Direct prompting} & Mistral-7B-Instruct & 0.67 & 0.40 & - & - & - & - \\
     &  & GPT-4o-mini & 0.68 & 0.44 & - & - & - & - \\
     &  & Llama-3-8B-Instruct & 0.68 & 0.45 & - & - & - & - \\
     %\rowcolor{Gray}
     &  & \cellcolor{Gray}Ensemble (above 3) & \cellcolor{Gray}0.68 & \cellcolor{Gray}0.43 & \cellcolor{Gray}- & \cellcolor{Gray}- & \cellcolor{Gray}- & \cellcolor{Gray}- \\
    \cline{2-9}
     & \multirow{4}{*}{$S_2$: Fine-grained} & Mistral-7B-Instruct & 0.67 & 0.40 & 0.55 & 0.58 & - & - \\
     &  & GPT-4o-mini & 0.71 & 0.53 & 0.56 & 0.61 & - & - \\
     &  & Llama-3-8B-Instruct & 0.68 & 0.44 & 0.55 & 0.58 & - & - \\
     %\rowcolor{Gray}
     &  & \cellcolor{Gray}\cellcolor{Gray}Ensemble (above 3) & \cellcolor{Gray}0.68 & \cellcolor{Gray}0.44 & \cellcolor{Gray}0.59 & \cellcolor{Gray}0.61 & \cellcolor{Gray}- & \cellcolor{Gray}- \\
    \cline{2-9}
     & \multirow{4}{*}{$S_3$: Chain-Of-Thought} & Mistral-7B-Instruct & 0.67 & 0.42 & 0.54 & 0.57 & 0.65 & 0.68 \\
     &  & GPT-4o-mini & \textbf{0.73} & 0.57 & 0.55 & 0.61 & \textbf{0.71} & \textbf{0.73} \\
     &  & Llama-3-8B-Instruct & 0.72 & \textbf{0.62} & 0.59 & 0.63 & 0.58 & 0.64 \\
     %\rowcolor{Gray}
     &  & \cellcolor{Gray}Ensemble (above 3) & \cellcolor{Gray}0.71 & \cellcolor{Gray}0.52 & \cellcolor{Gray}\textbf{0.63} & \cellcolor{Gray}\textbf{0.64} & \cellcolor{Gray}0.70 & \cellcolor{Gray}0.71 \\
    \hline
    \end{tabular}
\end{table*}

In this section, we evaluate the performance of multiple LLMs for dispute resolution with the three different prompting strategies. We choose 3 representative LLMs - Mistral-7B-Instruct, Meta-Llama-3-8B-Instruct, and GPT-4o-mini. The first two are open-source LLMs which we access using HuggingFace's API inference interface, whereas the third one is a closed-source LLM from OpenAI which we access using their API interface. The choice of LLMs is constrained by our hardware and budget constraints. However, the proposed approach is not specific to any particular LLM and given sufficient hardware any other larger models could be explored. %\footnote{GPT-4o is 100x more costlier than GPT-4o-mini.}. 
We use temperature setting of 0 throughout our experiments so as to ensure more consistent and repeatable responses from LLMs. This choice is based on the recommendation by Renze and Guven~\cite{renze2024effect} to set the sampling temperature to 0 for problem solving tasks to ensure maximum reproducibility without compromising accuracy.

%\subsection{Evaluating Structured Summary}
\subsection{Ground truth creation}\label{secGroundTruth}
As described in Section~\ref{secDRSystem}, we obtain three different types of output from LLMs -- stronger party prediction, demand-wise decisions, and argument-wise evaluation. In order to evaluate the performance of LLMs for these three types of output, we need ground truth (or gold-standard) output to compare against. We compare stronger party predictions for each dispute against the winning party identified in the structured summaries of the disputes (Section~\ref{secStructuredSummary}). 
For the other two tasks, we need to explicitly create ground truth for comparison. %For demand-wise decisions, the LLM is prompted (in $S_2$ and $S_3$) to give opinion for each individual demand identified in the structured summary. 
In order to create the ground truth for the demand-wise, we prompt the GPT-4o-mini model with the original dispute text (which contains the final decision on the dispute) along with the list of demands of a party (identified as part of the structured summary) to obtain the {\em true} decision for each demand, i.e., whether each demand was accepted or rejected in the final decision. %We use this information as ground truth labels for demand-wise decisions to compare the performance of LLM predictions for each demand. 
Similarly, we obtain ground truth labels for each argument of each party depending on whether that particular argument was considered favourably by the judge while arriving at the final decision. An important point to note here is that ground truth creation prompt contains the entire details of a dispute including the final decision and analysis/discussion leading to it. On the other hand, in {\bf DRAssist}, the resolution prompts ($S_1$, $S_2$, and $S_3$) do not contain any details about the final decision, as the LLM is expected to providing opinion on the final decision by ONLY relying on agreed facts, disagreement aspects, arguments, prior cases, and relevant statutes (as shown in Table~\ref{tabResolutionPrompt}).

\noindent\textbf{Matching of demands/arguments in resolution output with those in the ground truth.} LLMs often do not re-produce the list of demands/arguments provided in their input (from the structured summaries) verbatim in their resolution output. Hence, we can not use exact matching of the demands/arguments for comparison during evaluation. Their wordings may change or even their relative order changes in some rare cases. Hence, it is important that for each dispute, for each contesting party, the demands/arguments in the ground truth are {\em matched} with those in the resolution output. Here, we use cosine distance between their text embeddings\footnote{\url{https://huggingface.co/sentence-transformers/all-mpnet-base-v2}} to create a cost matrix and then use the linear sum assignment algorithm\footnote{\url{https://docs.scipy.org/doc/scipy/reference/generated/scipy.optimize.linear_sum_assignment.html}} to obtain an one-to-one mapping.

\subsection{Evaluation Metrics}
We evaluate the LLM output on the following three aspects (as described in Section~\ref{secDRSystem}) using two evaluation metrics.

%\begin{enumerate}
    %\item 
\noindent $\bullet$ {\bf Stronger party prediction}: For each dispute, we compare the \textit{predicted stronger party} with the \textit{gold-standard winning party}. This is applicable for all three prompting strategies - $S_1$, $S_2$, and $S_3$. We report {\em Accuracy} as well as {\em macro-averaged F1-score} (over two classes corresponding to the two contesting parties). In all the prompting strategies, along with stronger party prediction, corresponding {\em justification} is also generated. We compare the LLM-generated justifications with the gold-standard justifications using the following three metrics -- ROUGE-1, ROUGE-L\footnote{\url{https://pypi.org/project/rouge-score/}}, and BERTScore~\cite{zhang2019bertscore}.

    %\item 
\noindent $\bullet$ {\bf Demand-wise decisions}: %This is more fine-grained and hence more challenging aspect than the stronger party prediction. Here, 
For each dispute, for each demand by each contesting party, the LLM predictions (ACCEPTED or REJECTED) are compared with the gold-standard decisions for the corresponding demands. %This is only applicable in case of $S_2$ and $S_3$ prompting strategies. 
We report {\em Accuracy} as well as {\em macro-averaged F1-score} (over two classes -- ACCEPTED and REJECTED).

    %\item 
\noindent $\bullet$ {\bf Argument-wise evaluation}: This aspect evaluates how well the LLM is able to reason towards the stronger party and demand-wise decisions. Here, for each dispute, for each argument made by each contesting party, the LLM evaluations (STRONG or WEAK) are compared with the gold-standard evaluations of the corresponding arguments. We report {\em Accuracy} as well as {\em macro-averaged F1-score} (over two classes -- STRONG and WEAK).
%\end{enumerate}

\subsection{Baselines}
We consider two simple baselines for all the three types of LLM output described above. 
%\begin{enumerate}
    %\item 

\noindent $\bullet$ {\bf Majority Baseline}: Majority label in case of stronger party prediction is the contesting party which more often wins the dispute in the dataset, i.e., {\em insured party} in $D_{AI}$ and {\em complainant} in $D_{DN}$. For every dispute, this baseline predicts the majority label only. Similarly, for the other two tasks (demand-wise decisions and arguments evaluation), the majority label is found for a dataset and for each demand/argument, the same label is always predicted.

    %\item 
\noindent $\bullet$ {\bf Random Baseline}: As per this baseline, for any dispute, one of the contesting parties is \textit{randomly} chosen as the overall stronger party. Similarly, for the other two tasks, the random label (ACCEPTED/REJECTED or STRONG/WEAK) is chosen for each demand and argument.
%\end{enumerate}

\subsection{Ensemble}
For each prompting strategy, we are using 3 different LLMs. We explore a simple {\em ensemble} approach which combines predictions across these three LLMs by considering a majority vote for all the three types of predictions - stronger party prediction, demand-wise decisions, and argument-wise evaluation. For the later two, we need to establish a mapping between demands/arguments across the 3 LLM outputs which is achieved indirectly through their individual one-to-one mappings with the ground truth (see Section~\ref{secGroundTruth}). %As described in Section~\ref{secGroundTruth}, we already establish one-to-one mapping between demands/arguments across the ground truth and the individual LLM resolution output. 

%TODO: Evaluation of auto-generated structured summaries. Either compare the summaries with each other or compare each summary with a gold-standard summary.

%TODO: Re-generate the structured summaries by adding one more point -- Justification / rationale for the final decision.

%\subsection{Evaluating Resolution Performance of LLMs}
\subsection{Evaluation Results}
Table~\ref{tabResults} show the evaluation results for stronger party prediction, demand-wise decisions, and argument-wise evaluation, for both the datasets. % $D_{AI}$ and $D_{DN}$. 
%The results by all the three prompting strategies are shown in these tables which are compared against the two baselines. 
For stronger party prediction, for both $D_{AI}$ as well as $D_{DN}$, $S_3$ achieves the best overall macro-F1. For $D_{AI}$ there is a big improvement over the baselines, as $S_3$ (Ensemble) achieves macro-F1 of $0.78$ as compared to the baseline performance of $0.53$. On the other hand, for $D_{DN}$, $S_3$ (Llama) achieves macro-F1 of $0.62$ as compared to the baseline performance of $0.46$. If we observe relative performance among the three prompting strategies, it can be observed $S_3$ is better than $S_2$ and $S_2$ is better than $S_1$ for both the datasets. Hence, this establishes the effectiveness of chain-of-thought like reasoning (in terms of evaluation of arguments) before coming to the final conclusion. 

Predicting demand-wise decisions is a more difficult and fine-grained task as compared to merely predicting the stronger party. We observe the best macro-F1 of $0.64$ for $S_3$ in both the datasets, higher than the baseline performances of around $0.5$. We do not observe much difference in $S_2$ and $S_3$ performances in $D_{AI}$, though in $D_{DN}$, $S_3$ performs little better than $S_2$ for demand-wise decision predictions. In case of argument-wise evaluation, $S_3$ achieves macro-F1 of $0.6$ and $0.73$ for the datasets $D_{AI}$ and $D_{DN}$, respectively, which is much better than the baseline performance of around $0.5$. 

\begin{figure}[t]
\centering
\includegraphics[width=0.9\columnwidth,height=0.82\columnwidth]{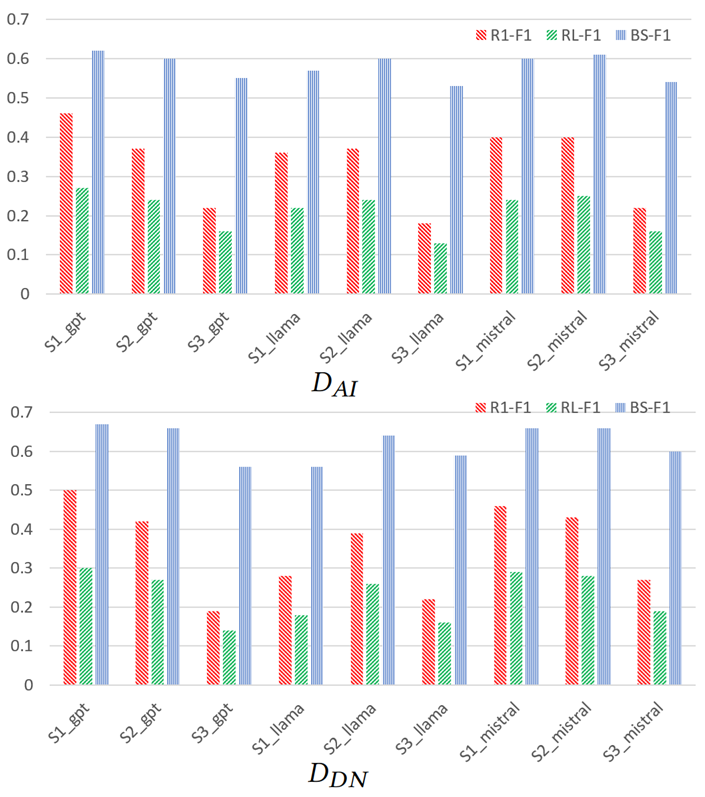}
%\caption{Evaluation of justifications generated by multiple LLMs with different prompting techniques (R1-F1: F1-score for ROUGE-1, RL-F1: F1-score for ROUGE-L, BS-F1: F1-score for BERTScore).} \label{figJustificationEval}
\caption{Evaluation of justifications generated by multiple LLMs with different prompting techniques.} \label{figJustificationEval}
\end{figure}

We also evaluate the quality of justifications generated by the LLMs for identifying a particular party as the stronger party. We compare the generated justifications with the ground truth justifications (obtained as part of the structured summaries) using ROUGE-1, ROUGE-L and BERTScore, averaged over multiple disputes. Figure~\ref{figJustificationEval} shows these evaluation results for various techniques, i.e., 9 combinations of 3 LLMs with 3 different prompting strategies, for both the datasets. Please note that for each technique, we consider only those disputes for averaging the evaluation metrics where the stronger party predictions are correct. This is to ensure that the evaluation only focuses on the justification and not on the prediction of stronger party. It can be observed that overall GPT and Mistral are better than Llama in generating the justifications for stronger party prediction. Also, there is another observation which is quite counter-intuitive. The performance of $S_3$ is worse than $S_1$ and $S_2$ even though its stronger party prediction is better. Upon investigation, it was found that several justifications generated by $S_3$ are mostly following a similar template where the justifications are of the form -- {\it Party A's arguments are stronger because... and party B's arguments are weaker.} This happens because $S_3$ first evaluates the arguments of both the parties before deciding the stronger party and hence the justification is mostly based on such comparison of relative strengths of the arguments. On the other hand, in $S_1$ and $S_2$, the LLM directly predicts the stronger party and produces a much better and elaborate justification.

%TODO: Evaluate quality of justification (for stronger party) by comparing it with justification obtained as part of structured summary.

%TODO: Resolution step to be run with more models - Mistral, SaulLM, Phi-3.5

%TODO: Add a random and majority baselines
% Winning Party	Count	insurance company	insured party	partial					
% insurance company	47	22	23	2				gpt-4o-mini	Majority baseline
% insured party	74	8	61	5				0.685950413	0.611570248
% partial	9	1	8	0					

% Winning Party	Count	insurance company	insured party	partial					
% insurance company	47	33	14					Random baseline	
% insured party	74	39	35					0.561983471	
% partial	9								

\subsection{Error Analysis}
Overall, the performance of stronger party prediction in $D_{AI}$ is better as compared to $D_{DN}$. Also, unlike the ensemble in $D_{AI}$, the ensemble in $D_{DN}$ is not able to achieve better results than the individual LLMs. The main reason behind this was observed to be the below par performance of Mistral for $D_{DN}$ where it always predicts ``complainant'' as the stronger party in all but 2 disputes. This also results in pulling down the ensemble performance. Such bias against ``respondent'' is also observed in the other two LLMs but to a lesser extent. Even in the best performing setting ($S3$ with Llama), the recall for the ``respondent'' label is just $0.31$. Analyzing the root cause for such bias and exploring ways to overcome the bias is important, which we leave for the future work.% by altering the resolution prompt accordingly.

We also analyzed the resolution output in case of our best performing $S_3$ strategy in detail for the error cases, i.e., the disputes for which the predicted stronger party is other than the ground truth winning party. Broadly, two sources of errors were found. First source is \textbf{incorrect structured summary generation} where errors in the structured summarization stage propagate to the resolution stage. See one example of such error in Table~\ref{tabErrorAnalysis} where one of the argument captured for the insurance company is {\em fallacious} or {\em self-harming}. This leads to incorrectly identifying this argument as WEAK, leading to the insurance company not getting identified as the stronger party. Another source of error is \textbf{incorrect reasoning}. See the second example in Table~\ref{tabErrorAnalysis} where the agreed facts (shown in blue) clearly mention that the vehicle was submerged in water. Here, GPT-4o-mini incorrectly identifies the insured as the stronger party whereas Mistral-7B-Instruct correctly identified the insurance company as the stronger party. See the difference in the reasoning by these two LLMs for one of the most important arguments by the insurance company (shown in red for GPT and in green for Mistral). Here, the reasoning provided by Mistral is clearly more consistent and logical in the light of the agreed facts, as compared to the one provided by GPT. In future, we plan to explore multi-agent frameworks where multiple LLM agents can collaborate to produce a better resolution output.%evaluation of arguments (and also a better final resolution).

\begin{table}[]\footnotesize
    \centering
    \caption{Error analysis}
    \label{tabErrorAnalysis}
    \begin{tabular}{p{0.97\columnwidth}}
    \hline
    \textbf{Example of incorrect structured summary:}\\
    \hline
    %0_0_RP_2221_2019.txt
    {\it \textbf{Arguments of the insurance company}:\newline
{\color{red}1. The driver, Riyaz Mohammed, possessed a valid and effective Driving License at the time of the accident, and the Insurance Company was informed of this in writing during the submission of documents}: \textbf{WEAK} : The insurance company claims that the insured party did not provide the driving license, but it does not provide evidence to support its assertion that the insured party failed to inform them in writing about the driver's valid license.}\\
    \hline
    \textbf{Example of incorrect reasoning:}\\
    \hline
    %0_0_FA_229_2015.txt
    {\color{blue}{\it \textbf{Facts agreed by both parties}:}\newline
    ...\newline
    {\it On 29.07.2010, the vehicle was submerged in floodwater, resulting in significant damage and the engine ceasing to function.}}\newline
    ...\newline
    {\it \textbf{Arguments of the insurance company}:}\newline
    ...\newline
    {\it 1. The insured party violated the terms and conditions of the insurance policy by failing to ensure proper precautions to safeguard the vehicle at all times: WEAK : {\color{red}The argument lacks specific evidence demonstrating how the insured failed to take precautions, making it less compelling.}} (with GPT-4o-mini)\\
    {\it 1. The insured party violated the terms and conditions of the insurance policy by failing to ensure proper precautions to safeguard the vehicle at all times: STRONG : {\color{teal}The policy terms and conditions explicitly state that the insured party is required to take proper precautions to safeguard the vehicle. The vehicle was submerged in floodwater, which could have been prevented with proper precautions.}} (with Mistral-7B-Instruct)\\
    \hline
    \end{tabular}

\end{table}

%TODO: Error analysis of incorrect decision prediction
% 0_0_FA_1584_2018.txt
% ANALYSIS	Counsel for the insurance company has failed to bring to my notice any rule or regulation under the Insurance Act which requires the insured to first get the damaged vehicle repaired by spending money from his own pocket and then it would be redeemed by the insurance company .
% ANALYSIS	The very purpose of buying the insurance policy is that the insured is not required to spend money in case of any damage or loss of his insured vehicle .
% ANALYSIS	If the argument of the insurance company is accepted , the very purpose of buying the insurance policy would be negated .

% Justification by GPT: 2. **Policy Terms**: While the specific terms of the policy are not provided, it is standard practice in vehicle insurance that the insured must repair the vehicle and provide proof of repairs to claim the assessed amount for repairable damages. The insurance company’s position that the complainant must first repair the vehicle aligns with common insurance practices.

%Additional refinement:
% 1. One-step resolution vs multi-step resolution (chain of legal reasoning)
% 2. User provides counter-argument or counter-justification for predicted decision. Will chat gpt change its prediction?

\section{Conclusions and Future Work}
In this paper, we presented {\bf DRAssist}, a prototype LLM-based Dispute Resolution Assistance system which works in zero-shot manner. We focused on two different domains of disputes - (i) automobile insurance disputes and (ii) domain name disputes. We created dispute datasets from these two domains and also enriched these datasets with structured summarization of each dispute and ground truth resolution output at various levels such as winning party, demand-wise decisions and argument-wise evaluation. We explored three different prompting strategies using three LLMs from different families (GPT, Mistral and Llama) in DRAssist. %, for obtaining the resolution assistance for each dispute. 
We observed that for identification of the stronger party, the prompting strategy $S_3$ which is based on the principle of Chain-of-Thought (CoT) performed better than others where the LLM is first made to analyze and evaluate each individual argument of both the contesting parties before deciding on the overall stronger party between them. For other fine-grained predictions of demand-wise decisions and argument-wise evaluations, though $S_2$ and $S_3$ perform better than the baselines, there is still some scope of improvement which we would pursue as future work. 
For the domain names dispute dataset, we observed that all the LLMs %(especially Mistral) 
are biased towards predicting complainant as the stronger party. We plan to analyse this phenomenon further and improve our approach to address such bias. We also plan to conduct an user study to assess the qualitative impact of our DRAssist system for resolving a dispute. Here, the most important user category would be the human expert (e.g., judge, arbitrator, mediator) whose responsibility is to resolve a dispute. Also, another category of the users would be the contesting parties who would be interested to see the evaluation of their arguments (weak or strong) and overall strength of their case.

\bibliographystyle{ACM-Reference-Format}
\bibliography{ref}





% LNCS Homepage, \url{http://www.springer.com/lncs}, last accessed 2023/10/25
% \end{thebibliography}

% \begin{appendices}

% \end{appendices}
\end{document}